\documentclass[letterpaper, 10 pt, conference]{ieeeconf}  

\IEEEoverridecommandlockouts                              

\usepackage[dvipsnames]{xcolor}

\usepackage{graphicx}
\usepackage{amsmath,amssymb}
\usepackage{bm}

\newcommand{\argmin}{\mathop{\rm arg~min}\limits}
\newcommand{\textbfb}[1]{\color{red}\textbf{#1}\color{black}}
\newcommand{\textb}[1]{\color{blue}#1\color{black}}

\usepackage{booktabs}
\usepackage{tabularx}
\usepackage{multirow}
\usepackage{verbatim}
\usepackage{subcaption}

\usepackage{cite} 
\usepackage{hyperref} 
\hypersetup{%
 colorlinks=true%
 }

\title{\LARGE \bf
REF$^2$-NeRF: Reflection and Refraction aware Neural Radiance Field
}

\author{Wooseok Kim$^{1}$, Taiki Fukiage$^{2}$ and Takeshi Oishi$^{1}$
\thanks{$^{1}$Wooseok Kim and Takeshi Oishi are with Institute of Industrial Science, The University of Tokyo, 153-8505, Tokyo, Japan
        {\tt\small \{kim, oishi\}@cvl.iis.u-tokyo.ac.jp}}%
\thanks{$^{2}$Taiki Fukiage is with NTT Communication Science Laboratories, Kanagawa, Japan {\tt\small taiki.fukiage@ntt.com}}%
}

\begin{document}

\maketitle
\thispagestyle{empty}
\pagestyle{empty}

\begin{abstract}

Recently, significant progress has been made in the study of methods for 3D reconstruction from multiple images using implicit neural representations, exemplified by the neural radiance field (NeRF) method. Such methods, which are based on volume rendering, can model various light phenomena, and various extended methods have been proposed to accommodate different scenes and situations. However, when handling scenes with multiple glass objects, e.g., objects in a glass showcase, modeling the target scene accurately has been challenging due to the presence of multiple reflection and refraction effects. Thus, this paper proposes a NeRF-based modeling method for scenes containing a glass case. In the proposed method, refraction and reflection are modeled using elements that are dependent and independent of the viewer's perspective. 
This approach allows us to estimate the surfaces where refraction occurs, i.e., glass surfaces, and enables the separation and modeling of both direct and reflected light components. 
The proposed method requires predetermined camera poses, but accurately estimating these poses in scenes with glass objects is difficult. 
Therefore, we used a robotic arm with an attached camera to acquire images with known poses.
Compared to existing methods, the proposed method enables more accurate modeling of both glass refraction and the overall scene.
\end{abstract}

\section{Introduction}


3D reconstruction from 2D images is a well-established technique; however, there is still room for improvement in terms of modeling scenes that involve transparent objects, e.g., glass. 
Glass is commonly used in the real world; thus, there is a great demand for modeling scenes containing transparent objects~\cite{klank2011transparent, sajjan2020clear, fang2022transcg, jiang2022a4t, tang2021depthgrasp}.
Unfortunately, common image sensors cannot observe transparent objects directly, and such objects produce various effects, e.g., reflections and refractions, which prevent conventional photogrammetry methods from modeling such scenes correctly.

Recent advances in implicit neural representations~\cite{nerf} of 3D scenes have made it possible to model and synthesize novel views of scenes including reflection and refraction effects~\cite{IchnowskiAvigal2021DexNeRF}. 
Prior to the introduction of the neural radiance field (NeRF) technique, researchers studied these photometric effects as physical occurrences and developed techniques to replicate scenes based on the physical principles. 
However, modeling real-world scenes based on physical models from images is an ill-posed problem that requires various constraints. 
In contrast, NeRF-based methods allow neural networks to learn these complex phenomena to synthesize new views effectively and enable geometry modeling. 
The ability of these methods to model transparent objects, metallic objects, objects in transparent medias, and objects in liquids without the need for complex models or constraints was a major development in the field. 

\begin{figure}[t]
\begin{center}
\includegraphics[width=0.87\linewidth]{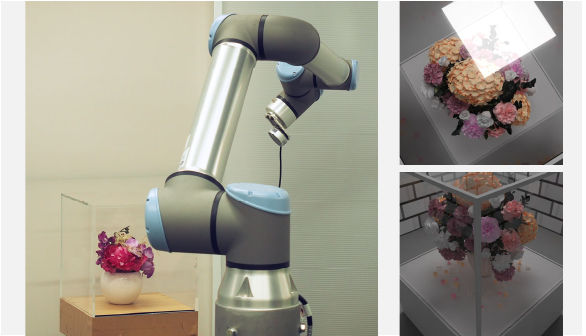}
\caption[]{
Image acquisition system and example images of a scene including a glass case and objects. Images of those scenes contain effects of light ray reflection and refraction, which vary depending on viewpoint.}
\label{fig:scanning}
\end{center}
\end{figure}

However, even with these techniques, modeling scenes with multiple transparent surfaces remains a challenge. 
Scenes frequently feature glass objects, as demonstrated by the showcase depicted in Fig.~\ref{fig:scanning}.
When such objects appear in a scene, they generate multiple reflections and refractions along the viewer's line of sight, which increases the task complexity of generating neural fields that capture the corresponding objects accurately. 
In such scenes, camera tracking is also difficult, making it even more challenging to refine camera poses while training NeRF. 
%

Thus, in this paper, we propose a neural modeling method that considers the characteristics of scenes containing multiple glass surfaces, particularly objects enclosed in a glass case. 
The proposed method introduces two networks to handle refraction and reflection independently, and it decomposes the view-independent and view-dependent components of these effects.
In the case of refraction, the view-independent component is the refraction point, and the view-dependent component is modification of the ray's direction. 
Accordingly, the proposed method learns to synthesize new viewpoint images and estimates the position and magnitude of the refraction within the given scene. 
An additional network decomposes the direct and reflection components in geometric and photometric aspects.  

The contributions of this paper can be summarised as follows:
\begin{itemize}
    \item 
We propose a neural network-based approach for modeling scenes with multiple glass surfaces, focusing particularly on objects inside a glass case. 
This approach involves using two separate networks to handle the effects of refraction and reflection.
\item We introduce a framework that handles refraction and reflection efficiently by learning the view-dependent and view-independent components separately.
\item The proposed method decomposes direct and reflection components in geometric and photometric terms and estimates refraction position and magnitude in the scene.
\item  We demonstrate that our method works for simulation datasets and real scenes captured by a robotic arm-based measurement system.
\end{itemize}

\section{Related work}

In the following, we briefly review previous studies related to the multiview reconstruction and learning of scenes that include transparent objects, e.g., glass. 

\subsection{Multi-view method for transparent object}
Common image-based 3D reconstruction methods based on the structure from motion and multiview stereo (MVS)~\cite{Seitz2006}\cite{furukawa2009accurate}\cite{colmap}\cite{meshroom} techniques estimate the surface geometry using triangulation from feature points or patches. 
MVS technology has matured recently, and its use is common in various practical applications. 
In addition, the development of deep learning has made them more robust and accurate.
However, strong reflection affects feature and texture matching, and refraction distorts the estimated surface shape, which makes it difficult to reconstruct such scenes using conventional MVS methods.
%
%
%

Methods have also been proposed to model transparent objects and objects behind or within them. 
For example, several methods operate under a controlled environment, e.g., using background patterns~\cite{Lyu2020}\cite{wu2018full} and polarization~\cite{miyazaki2004}\cite{NEURIPS2019Lyu}. 
However, these methods attempt to estimate the surface shape of the target transparent object; thus, they are unsuitable for modeling regions or objects behind transparent objects.
A popular scene where it is difficult to control the environment and target behind or within transparent objects is modeling underwater environments.
For underwater scenes, the refraction of light rays occurs at the surface between the lens and the water~\cite{qiao2019underwater, beall20103d, jordt2016refractive, Chadebecq2017}. 
In other words, multiple reflections and refractions are not considered or require prior ray calibration to model more complex scenes. 

Some methods employ supervised learning approaches~\cite{li2020through}; however, the results are dependent on the available training data, and such methods do not handle complex light effects as well as other model-based methods.




\subsection{Neural Radiance Fields (NeRF)}
The NeRF technique~\cite{nerf} can represent various optical phenomena because the trained neural network data, density, and color fields are based on volume rendering~\cite{volumerendering}. 
NeRF and its variants optimize the fields represented by implicit neural functions to reduce the similarity error between the input and rendered images. 
A well-trained NeRF model allows us to reconstruct 3D scenes or synthesize novel views from the neural models. 
However, the original NeRF has some drawbacks; thus, various methods have been proposed to improve NeRF in terms of acceleration~\cite{instantngp, plenoxels, yu2021plenoctrees}, synthesized image quality enhancement~\cite{mipnerf, mipnerf360, wang2022nerf}, and robustness against various conditions~\cite{yu2021pixelnerf,barf,nerfw,refnerf,nerfies,wang2021nerf}. 
Originally, NeRF was designed to realize novel view synthesis; however, more accurate shape reconstruction is achievable using, for example, signed distance function (SDF)-based approaches~\cite{wang2021neus}\cite{neus2}. 
There are also various approaches to efficiently take images and reconstruct 3D scenes with the implicit neural representations using ground-based or aerial robots\cite{zeng2023efficient, lee2022uncertainty, jin2023neu, ran2023neurar, yan2023active, zhou2023inf}.

NeRF has shown promising results; however, similar to common 3D reconstruction methods, it assumes a straight light ray and generates poor results when refraction is present in the input images. 
%
%
In addition, reflections in a scene appear as if another scene exists behind the observable reflective and transparent objects, and those reflections may not be visible depending on the viewpoint from which the scene is observed. Note that this phenomenon contradicts the view-consistency assumption of NeRF.

\subsection{NeRF for reflective scenes}
Several NeRF variants have been proposed to handle scenes that include reflective surfaces. 
For example, NeRFReN~\cite{nerfren} handles reflections by assuming reflective plane surfaces and separating the scene into transmitted and reflected components utilizing two rendering paths.
In addition, NeuS-HSR~\cite{qiu2023looking} also estimates an auxiliary plane that separates the reflection components, which facilitate reconstruction of an object inside a glass case. 
The Neural Transmitted Radiance Fields method~\cite{NEURIPS2022_transmitted} detects recurring edges in the input images to optimize transmission and reflection components independently. 

The MS-NeRF~\cite{msnerf} technique separates the input scene into multiple spaces and estimates density fields and feature fields for these spaces. 
Here, the feature fields reduce the number of estimated parameters while estimating the color map and weights of the spaces. 
%
The neural point catacaustics method~\cite{Kopanas2022} introduces a pointwise neural warp field that represents the reflection from curved surfaces, which makes it possible to render the reflected points that are separated from the primary point cloud. 

The aforementioned methods can model scenes containing reflective objects effectively; however, they do not consider refraction by transparent objects, and they do not handle scenes containing multiple transparent objects.


\subsection{NeRF for refractive scenes}
The LB-NeRF method~\cite{lbnerf} addresses scenes in which objects are present inside a refractive medium. 
The LB-NeRF method handles refraction by simplifying it as an offset from a straight light ray. 
By adding the offsets to each sampled point's position prior to training NeRF's MLP, the LB-NeRF method models canonical space without refraction effects. 
Other methods based on the physical properties of reflective and refractive medium~\cite{samplingnerf,neref,Tong_2023_CVPR,Zhan_2023_ICCV} require additional information, e.g., a known image pattern, the refractive index, or a mask image of the refraction, as a clue to detect and estimate the refraction present in an image.

A number of NeRF-based methods have been proposed for reflective or transparent objects~\cite{IchnowskiAvigal2021DexNeRF}\cite{evonerf}\cite{dai2023graspnerf}\cite{refnerf}; however, it is difficult to apply these methods to the target scene considered in this study for the reasons described above.

\section{Preliminary and overview}

\begin{figure}[t]
    \centering
    \includegraphics[width=0.9\linewidth]{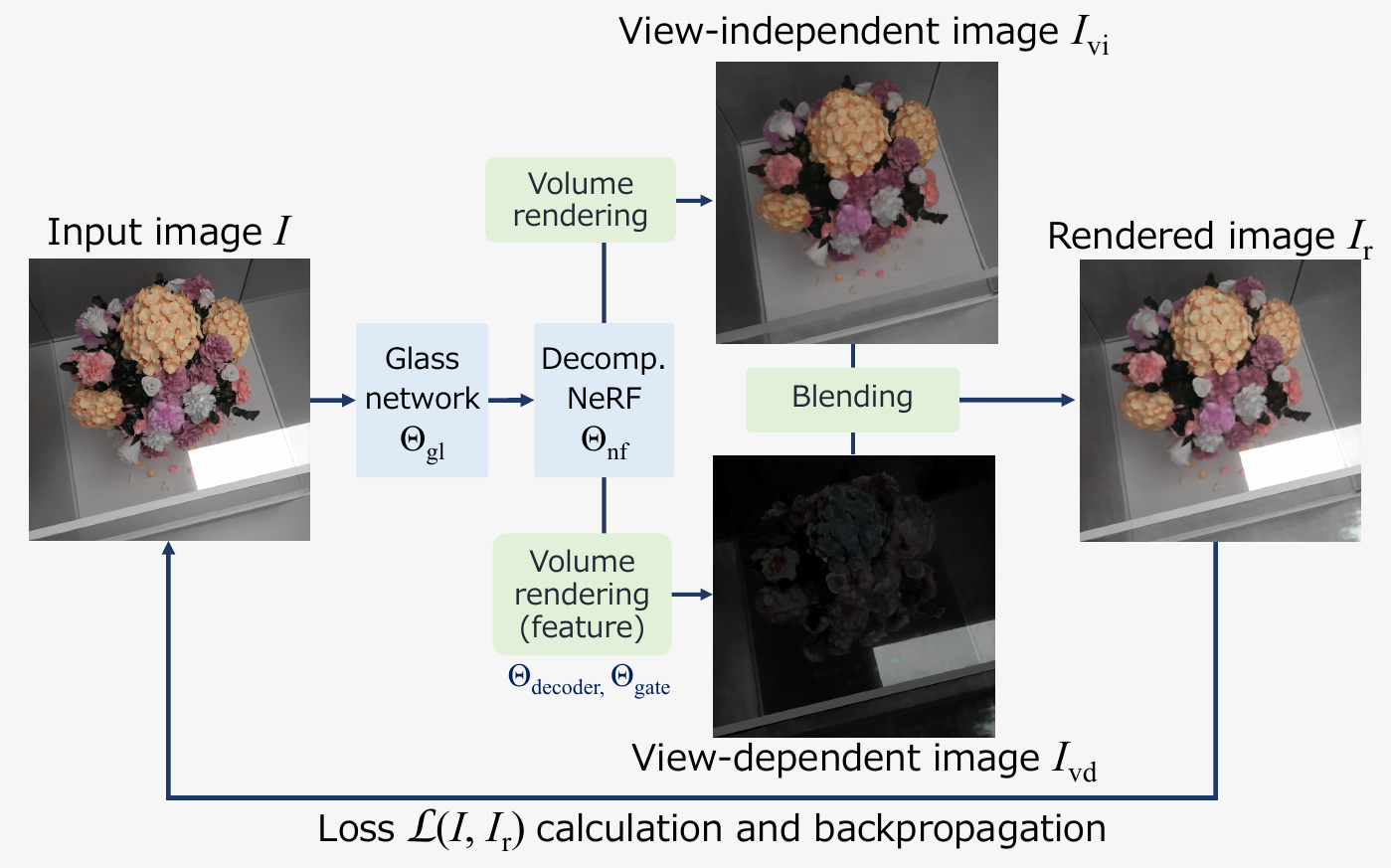}
    \caption{Overview of the proposed framework. Glass network MLP models refraction occurred by transparent object and adjusted each sampled position. Then, we decompose the scene into view-dependent and view-independent components to separate reflection from input images and model both. }
    \label{fig:overview}
\end{figure}

In this section, we summarize the refraction and reflection effects and provide an overview of the proposed framework, which is based on these effects.


\subsection{Refraction and reflection effects} 
\textit{Refraction} is a phenomenon whereby the direction of light changes due to differences in the refractive index between mediums. 
Note that refraction also changes according to the incident angle, and it follows Snell's law. 
Thus, the glass surface where refraction occurs is independent of the viewpoint; however, the amount of refraction depends on the viewpoint.
In addition, when light passes through a glass plate, it passes through the parallel boundary in the air-to-glass and glass-to-air order; thus, the ray is parallel to the original ray and is shifted according to the incident angle and the thickness of the glass. 

\textit{Reflection} occurs on the surface of the glass and, similar to refraction, is a phenomenon whereby the path of the light changes. 
Here, the reflected light intensity is largely distributed in the direction opposite to the incident angle relative to the surface normal, which means that the reflected light intensity is strongly dependent on the viewpoint.
Reflections, like mirrors, create a mirror object, i.e., it appears as if the same object is behind the glass. 
However, it differs from a mirror in that the mirror object is semitransparent because some of the light is reflected on the glass surface, and the remaining light is transmitted into the glass.
In other words, in addition to the view-independent objects, view-dependent semitransparent objects can be assumed to exist in the scene.

\subsection{Overview of proposed framework}
Based on the above considerations, the proposed framework is designed to estimate the refraction effect and separate the view-independent and view-dependent components present in the given scene. 
Figure \ref{fig:overview} shows an overview of the proposed method. 
In the proposed method, we assume that the input is multiple images $\{I_k\}(k=1,2,...,n)$ taken from different viewpoints, similar to existing NeRF variants.
Here, $n$ is the number of input images, and we omit the index $k$ in this section for simplicity. 

The proposed framework primarily comprises two independent MLPs, i.e., $\Theta_\mathrm{gl}$ and $\Theta_\mathrm{nf}$, to handle the refractive and reflective components independently, respectively. 
The former is referred to as the glass network. The glass network represents the refraction points and the amount of refraction, which give the parallel shift of the ray for sampling points in the latter network. 
The latter is the main NeRF network, which decomposes the direct and reflection components. 
Here, $\Theta_\mathrm{nf}$ represents the fields to render an image $I_\mathrm{vi}$ of a direct component and an image $I_\mathrm{vd}$ of a reflection component. 
The density and feature fields of the view-dependent component generate the image through the decoder and gate MLPs: $\Theta_\mathrm{dc}$, $\Theta_\mathrm{gt}$ \cite{msnerf}. 
The blended image of $I_\mathrm{vi}$ and $I_\mathrm{vd}$ is the output image $I_\mathrm{r} = I_\mathrm{vi} \oplus \alpha I_\mathrm{vd}$, where $\alpha$ is the blending parameter derived from the network.

The training process optimizes both networks while minimizing the loss $\mathcal{L}$ between the input and rendered images as follows: 
%
%
%
\begin{align}
\label{eq:entire_loss} 
\hat{\Theta}_\mathrm{gl}, \hat{\Theta}_\mathrm{nf}, \hat{\Theta}_\mathrm{dc}, \hat{\Theta}_\mathrm{gt} = \argmin_{\Theta_\mathrm{gl}, \Theta_\mathrm{nf}, \Theta_\mathrm{dc}, \Theta_\mathrm{gt}} \mathcal{L}(I, I_\mathrm{r}).
\end{align}

\section{Proposed framework}
\begin{figure*}[t]
    \centering
    \includegraphics[width=0.9\linewidth]
{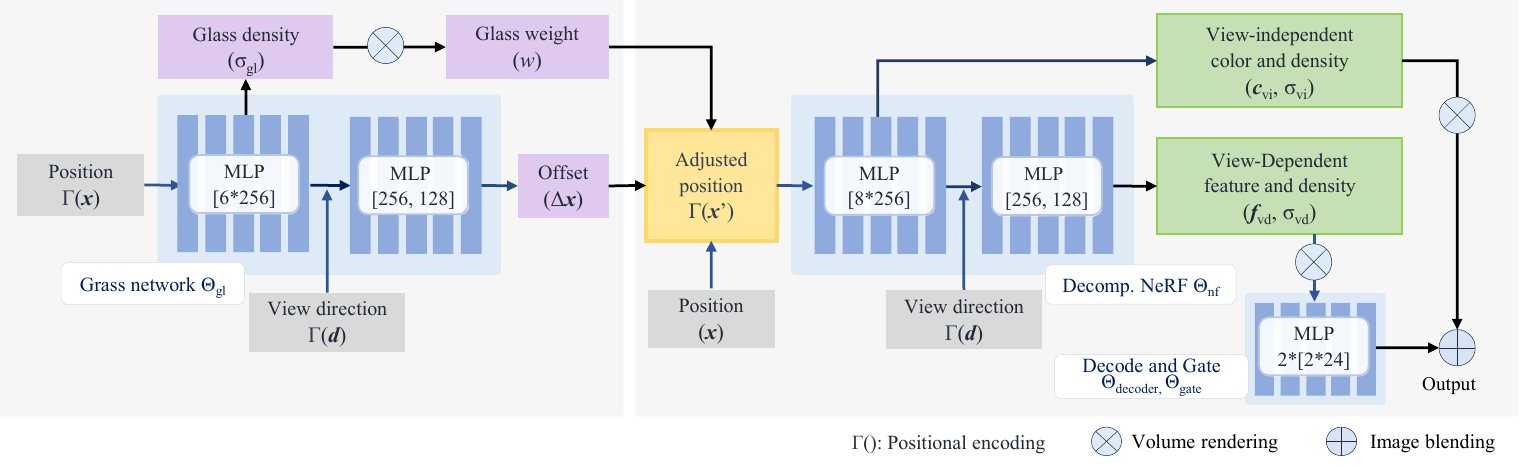}
    \caption[]{Network architecture of the proposed method. The glass network outputs the glass density and offset, which modify the ray by the refraction effect through glass walls. Here, the glass density is view-independent, and the offset is view-dependent. The NeRF network takes the adjusted position as input and outputs the view-independent and view-dependent densities and color or feature. The feature renderer provides the corresponding feature map, and the decoder and gate MLPs convert the rendered feature map to a view-dependent image with a blending weight. Finally, the image blending module generates the image by composing the rendered view-independent and view-dependent images. The training process minimizes the loss calculated from the composed image and the input image while optimizing the MLPs. 
    }
    \label{fig:network}
\end{figure*}

This section describes the proposed framework and its implementation in detail. 
Figure \ref{fig:network} shows the network architecture of the proposed framework and the rendering flow.



\subsection{Glass network}



%
%
\begin{figure}[t]
    \centering
    \includegraphics[width=0.9\linewidth]{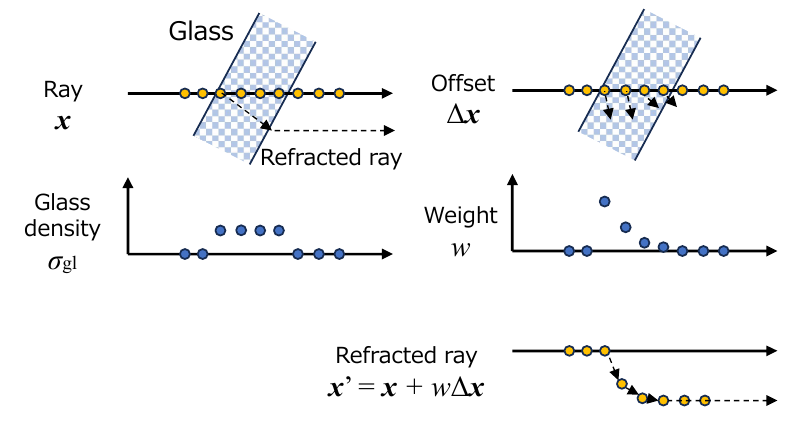}
    \caption{Structure of the proposed method to express light refraction as volume rendering using glass density to estimate the offset. Here, refraction is simplified as a parallel translation in 3D space occurring on the glass surface. We estimate the path of the light considering refraction by accumulating the vectors of this translation.}
    \label{fig:refraction_image}
\end{figure}
\begin{figure}[t]
    \centering
    \includegraphics[width=0.92\linewidth]{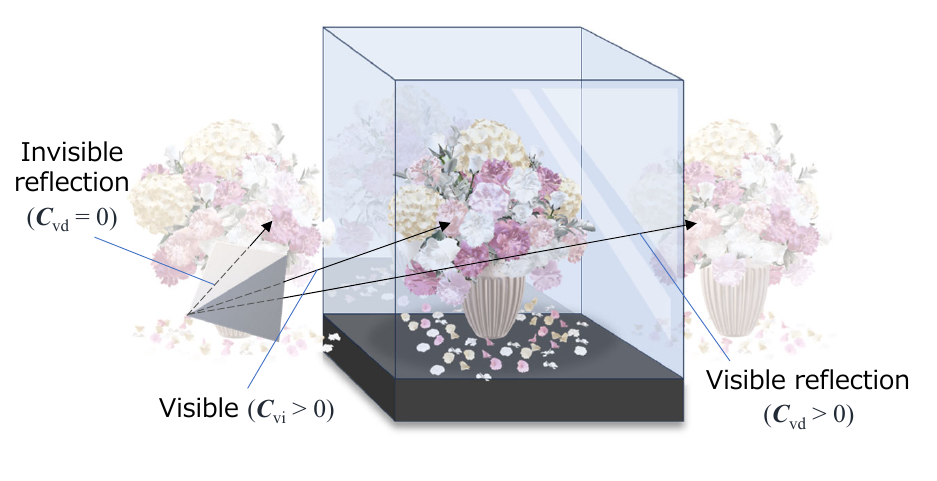}
    \caption{A proposal method structure that divides the scene into two fields. Elements that do not change depending on the viewpoint, e.g., objects and backgrounds in the scene, are represented in the view-independent field. Elements that do change depending on the viewpoint, e.g., reflections caused by glass and reflections from light sources, are represented in the view-dependent field, where the density changes based on the viewpoint.}
    \label{view_dependent}
\end{figure}

The glass network is employed to estimate the location of the glass surface where the refraction occurs and the amount of refraction.
The method used to simplify and express the refraction as an offset is similar that utilized in the LB-NeRF technique~\cite{lbnerf}. 
In contrast to LB-NeRF, the proposed method introduces the view-independent density field and view-dependent offset field to estimate the refraction surfaces and the offsets simultaneously.

Here, for a point $\boldsymbol{x}_i \in \mathbb{R}^3, (i = 1,2,...,N)$ sampled along a ray $\boldsymbol{r}$, the glass network $\Theta_\mathrm{gl}$ estimates the glass density $\sigma_\mathrm{gl}$, which indicates the degree to which that point is involved in the refraction. 
$N$ is the number of sampled points. 
$\Theta_{\mathrm{gl}}$ also outputs the offset vector $\Delta \boldsymbol{x}_i \in \mathbb{R}^3$, which represents the magnitude and direction of the refraction arising from the view direction $\boldsymbol{d}_i \in \mathbb{R}^3$ and position $\boldsymbol{x}_i$. 
In other words, the glass network takes the encoded $\boldsymbol{x}$ and $\boldsymbol{d}$ as inputs and outputs $\sigma_\mathrm{gl}$ and $\Delta \boldsymbol{x}$. This process is expressed as follows: 

\begin{equation}
\label{eq:glass_network}
\mathcal{F}_{\Theta_\mathbf{gl}} : \Gamma(\boldsymbol{x}),\Gamma(\boldsymbol{d}) \to \sigma_\mathrm{gl}(\boldsymbol{x}),\Delta\boldsymbol{x}(\boldsymbol{x},\boldsymbol{d}),
\end{equation}
where $\Gamma$ represents the positional encoding.

The sampling points are adjusted after refraction using the offset vectors, as shown in Fig. \ref{fig:refraction_image}.
Similar to NeRF's volume rendering~\cite{volumerendering}, in the proposed method, the refraction weight of each sampling point is calculated using the glass density $\sigma_{\mathrm{gl},i}$ and the distance between adjacent sampling points $\delta_i$ as follows:
\begin{align}
\label{eq:volume_rendering_offset}
w_i=T_{i}(1-exp(-\sigma_{\mathrm{gl},i}\delta_{\mathrm{gl},i})),\\
\label{eq:volume_rendering_probability}
T_{i}=exp(-\sum_{j=1}^{i-1}\sigma_{\mathrm{gl},j}\delta_{\mathrm{gl},j}).
\end{align}
%
%
%
As a result, we obtain the amount of ray shifting that represents the distance each sampling point moves from its original coordinate by adding the weighted offset cumulatively along the ray, and we estimate the adjusted position of the sampling points $\boldsymbol{x}'$ as follows:
%
\begin{equation} \label{eq:volume_rendering_offset_2}
\boldsymbol{x}'_i=\boldsymbol{x}_i+\sum_{j=1}^{i} w_j \Delta\boldsymbol{x}_j.
\end{equation}
%

\subsection{Decomposition NeRF}


We assume that an input scene can be separated into the view-independent component, which does not change based on the viewpoint, and the view-dependent component, which does change based on the viewpoint, as shown in Fig.~\ref{fig:overview}.
We then define two NeRF-like fields representing each component.

\subsubsection{View independent NeRF}
The view-independent components are represented using density $ \sigma_\mathrm{vi} \in \mathbb{R}$ and color $\boldsymbol{c}_\mathrm{vi} \in \mathbb{R}^3$, similar to the conventional NeRF method. 
However, the view-independent components do not require the view direction; thus, in the proposed method, we only use a former part of $\Theta_\mathrm{nf}$, which takes the position $\boldsymbol{x}'$ as input.
By performing volume rendering with the $(\sigma_\mathrm{vi},\boldsymbol{c}_\mathrm{vi})$ of the points sampled on a ray $\boldsymbol{r}$, we obtain the view-independent color $\boldsymbol{C}_{\mathrm{vi}}$ of a pixel corresponding to that ray as follows: 
\begin{equation} \label{volume_rendering_viewindependent}
\boldsymbol{C}_{\mathrm{vi}}(\boldsymbol{r})=\sum_{i=1}^{N}T_{\mathrm{vi},i}(1-exp(-\sigma_{\mathrm{vi},i}\delta_{\mathrm{vi},i}))\boldsymbol{c}_{\mathrm{vi},i}.
\end{equation}
Note that the calculation of $T_{\mathrm{vi},i}$ is the same as given in Eq.~\ref{eq:volume_rendering_probability}.

\subsubsection{View dependent NeRF}
The view-dependent components are separated using the feature field approach~\cite{msnerf}. 
MS-NeRF introduced the feature field, which extracts multiple spaces explicitly as the density and feature fields, and it functions effectively for scenes with several mirrors.
However, multiple glass plates generate reciprocal reflections of objects and light sources; thus, representing all spaces individually with a number of spaces is a highly complex task. 

In the proposed method, we address this problem using a single view-dependent feature field in addition to the previously describe view-independent NeRF. 
The feature field is represented using $\Theta_\mathrm{nf}$, which estimates the view-dependent density $\sigma_\mathrm{vd} $ and the $\theta$-dimensional feature vector $\boldsymbol{f}_\mathrm{vd}$ from an adjusted position $\boldsymbol{x}'$ and view direction $\boldsymbol{d}$ as follows: 
%
\begin{align}
\nonumber
\mathcal{F}_{\Theta_\mathbf{nf}}
: 
\Gamma(\boldsymbol{x}'),\Gamma(\boldsymbol{d})
&\to\\
\sigma_\mathrm{vi}(\boldsymbol{x}'),&\boldsymbol{c}_\mathrm{vi}(\boldsymbol{x}'),
\sigma_\mathrm{vd}(\boldsymbol{x}',\boldsymbol{d}),
\boldsymbol{f}_\mathrm{vd}(\boldsymbol{x}',\boldsymbol{d}).
\label{eq:nerf_network}
\end{align}
We obtain the feature vector $\boldsymbol{F}_\mathrm{vd}$ corresponding to a ray $\boldsymbol{r}$ by volume rendering along the ray for $\sigma_\mathrm{vd}$ and $\boldsymbol{f}_\mathrm{vd}$, 
In addition, we estimate the color $\boldsymbol{C}_\mathrm{vd}$ using a decoder MLP $\Theta_\mathrm{dc}$ and determine the blending parameter $\alpha$ using a gate MLP $\Theta_\mathrm{gt}$~\cite{msnerf}. 
Here, $\boldsymbol{C}_\mathrm{vd}(\boldsymbol{r})$ represents the reflection component corresponding to the pixel of the ray.
\begin{align}
\label{volume_rendering_viewdependent}
\boldsymbol{F}_\mathrm{vd}(\boldsymbol{r})=&\sum_{i=1}^{N}T_{i}(1-exp(-\sigma_{\mathrm{vd},i}\delta_{\mathrm{vd},i}))\boldsymbol{f}_{i}, \\
{\mathcal{F}_{\Theta}}_\mathrm{dc}&:\boldsymbol{F}_\mathrm{vd}(\boldsymbol{r}) \to \boldsymbol{C}_\mathrm{vd}(\boldsymbol{r}),\\ 
{\mathcal{F}_{\Theta}}_\mathrm{gt}&:\boldsymbol{F}_\mathrm{vd}(\boldsymbol{r}) \to 
\alpha(\boldsymbol{r}).
\end{align}
%
%
%
%
\subsection{Optimization}
Finally, we find the color $\boldsymbol{C}(\boldsymbol{r})$ of the pixel corresponding to the ray as follows:
\begin{equation} \label{color} 
\boldsymbol{C}(\boldsymbol{r})=\boldsymbol{C}_\mathrm{vi}(\boldsymbol{r}) + \alpha(\boldsymbol{r}) \boldsymbol{C}_\mathrm{vd}(\boldsymbol{r}).
\end{equation}
We train the MLPs \{$\Theta_\mathrm{gl}$,$\Theta_\mathrm{nf}$, $\Theta_\mathrm{dc}$, and $\Theta_\mathrm{gt}$\} by evaluating the rendered pixel color $\boldsymbol{C}(\boldsymbol{r})$ with that of the input image $\bar{\boldsymbol{C}}(\boldsymbol{r})$. 
Here, we utilize the same loss function utilized in the conventional NeRF method, i.e., the summation of L2 distances, which is expressed as follows:
\begin{equation}
\label{eq:image_loss} 
\mathcal{L}_\mathrm{render}=\sum
_{\boldsymbol{r} \in \boldsymbol{R}}\|\bar{\boldsymbol{C}}(\boldsymbol{r})-\boldsymbol{C}(\boldsymbol{r})\|^2,
\end{equation}
where $\boldsymbol{R}$ is a batch of sampled rays. 
In addition, we introduce L2 regularization loss $\mathcal{L}_\mathrm{offset}$ to train the glass network in a stable manner. 
This regularization loss prevents the neural network from learning biased offset and parallelly shifted scenes. 
\begin{equation} \label{eq:offset_loss} 
\mathcal{L}_{\mathrm{offset}} = \sqrt{ \sum_{\boldsymbol{x},\boldsymbol{d}} (\Delta \boldsymbol{x})^2}
\end{equation}
The entire loss $\mathcal{L}$ is an integration of these two loss functions:
\begin{align}
\label{eq:entire_loss_experiment} 
\mathcal{L}=\mathcal{L}_\mathrm{render}+\epsilon\mathcal{L}_\mathrm{offset},
\end{align}
where $\epsilon$ is a small number, which was set to $\epsilon=10^{-5}$ in our experiments. 

\section{Experiment}



\begin{figure}[t]
    \centering
    \includegraphics[width=0.3\linewidth]{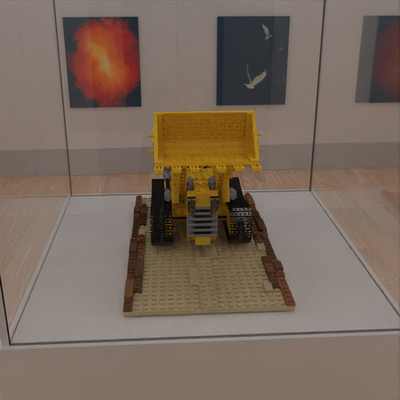}
    \includegraphics[width=0.3\linewidth]{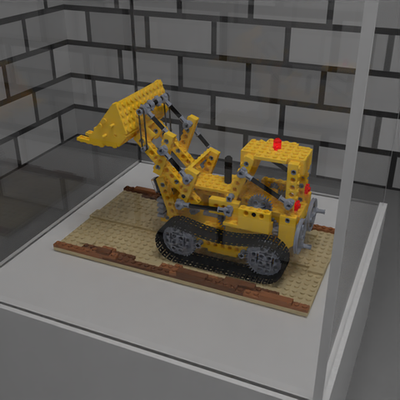}
    \includegraphics[width=0.3\linewidth]{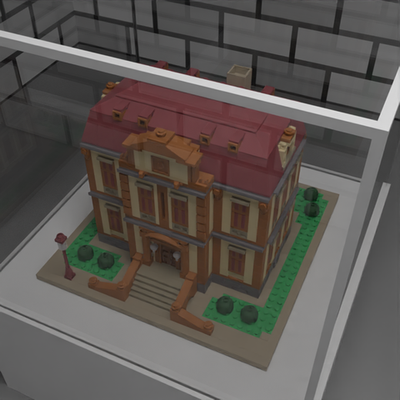}
\vspace{2pt}

    \includegraphics[width=0.3\linewidth]{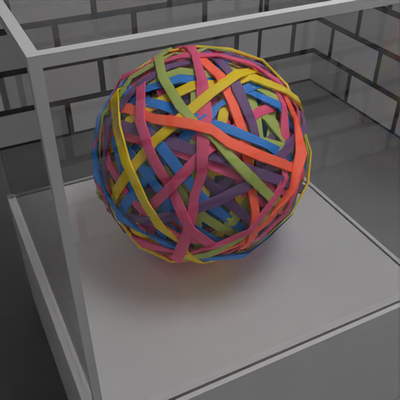}
    \includegraphics[width=0.3\linewidth]{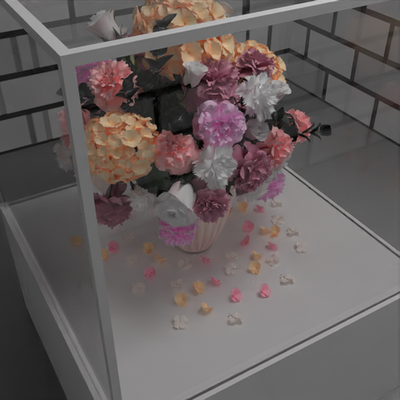}

    \caption{Example images of simulation dataset. Each image (from left-top to right-bottom) shows input images for Lego (Gallery), Lego, House, Color Ball, and Flower, respectively.}
    \label{fig:simulation_dataset}
\end{figure}
\begin{figure}[t]
    \centering
    \includegraphics[width=0.3\linewidth]{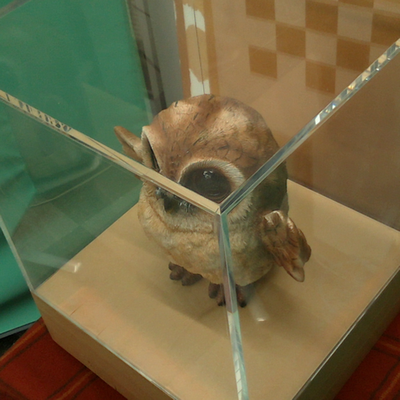}
    \includegraphics[width=0.3\linewidth]{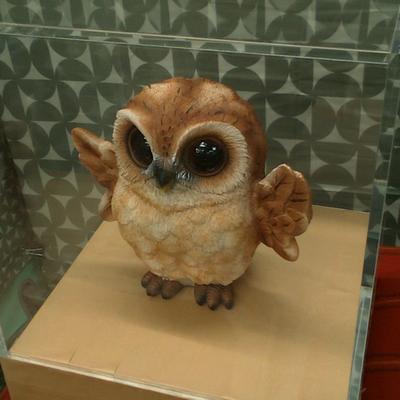}
    \includegraphics[width=0.3\linewidth]{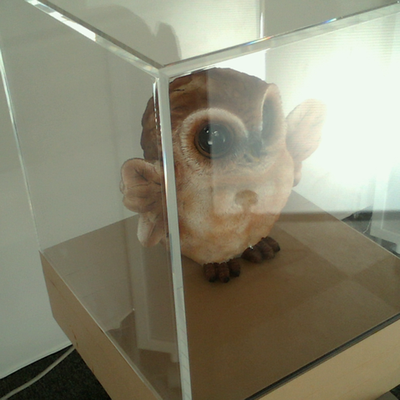}
\vspace{2pt}

    \includegraphics[width=0.3\linewidth]{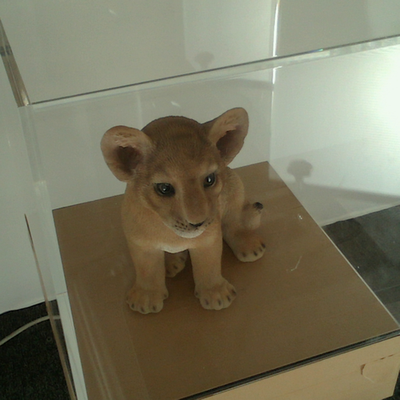}
    \includegraphics[width=0.3\linewidth]{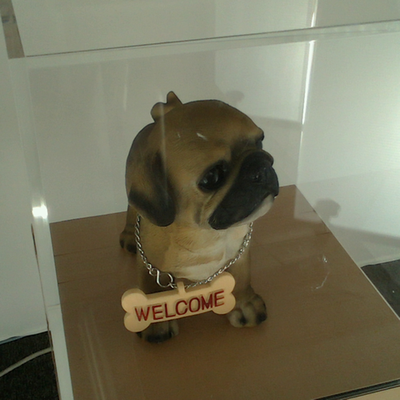}

    \caption{Example images of real-world dataset. Each image (from left-top to right-bottom) shows input images for Owl (simple bg), Owl (textured bg), Owl (white bg), Lion (white bg), Dog (white bg), respectively.}
    \label{fig:realworld_dataset}
\end{figure}

\begin{figure*}[t]
    \centering
    \includegraphics[width=0.16\linewidth]{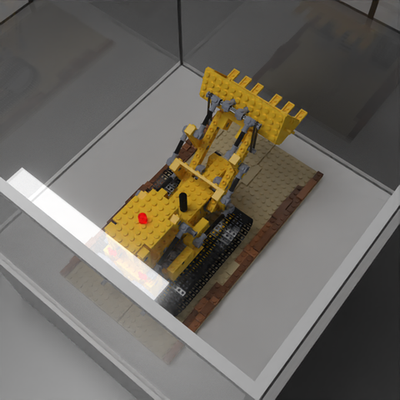} 
    \includegraphics[width=0.16\linewidth]{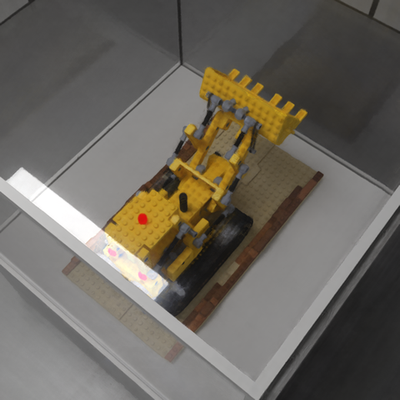} 
    \includegraphics[width=0.16\linewidth]{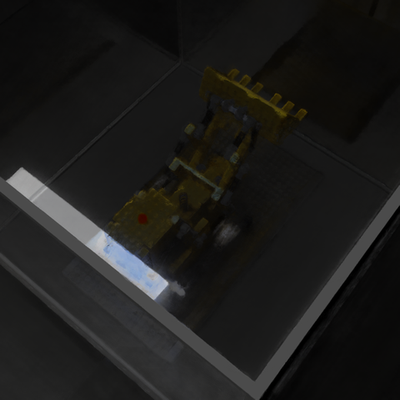} 
    \includegraphics[width=0.16\linewidth]{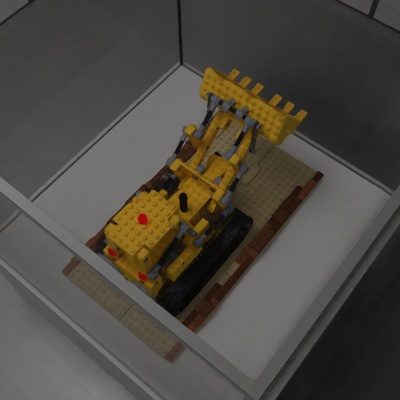} 
    \includegraphics[width=0.16\linewidth]{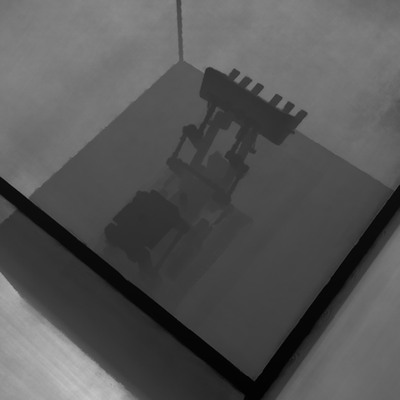} 
    \includegraphics[width=0.16\linewidth]{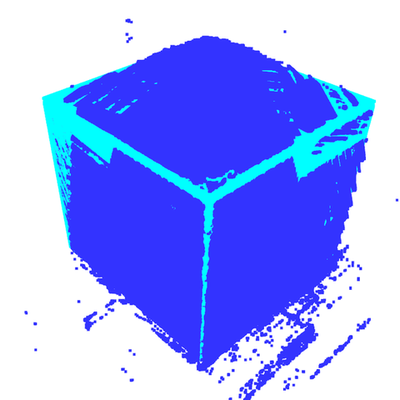}


%
%

\vspace{2pt}
    \includegraphics[width=0.16\linewidth]{fig/results_real/owl_009_gt.png}
    \includegraphics[width=0.16\linewidth]{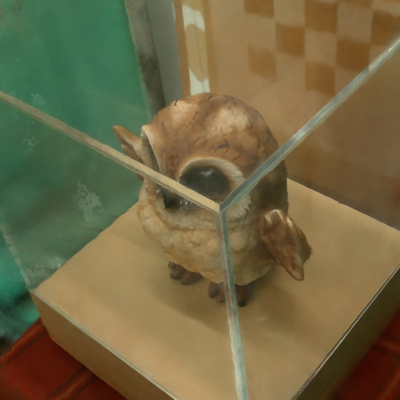}
    \includegraphics[width=0.16\linewidth]{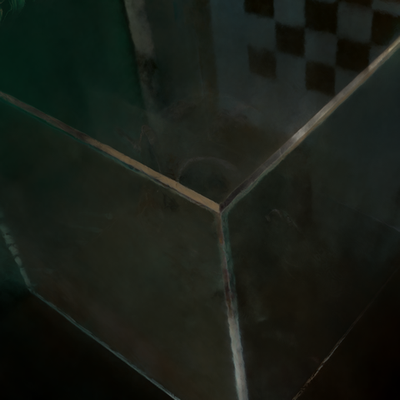}
    \includegraphics[width=0.16\linewidth]{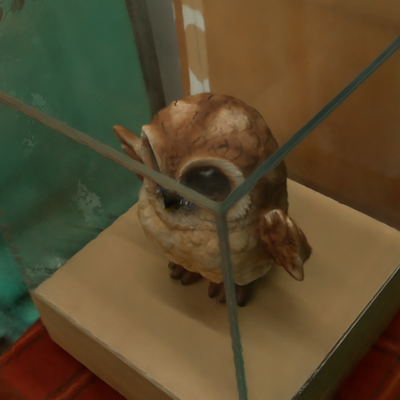}
    \includegraphics[width=0.16\linewidth]{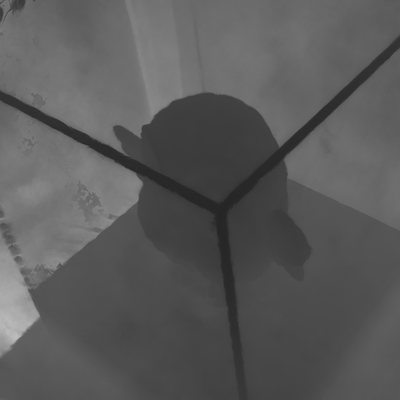}
    \includegraphics[width=0.16\linewidth]{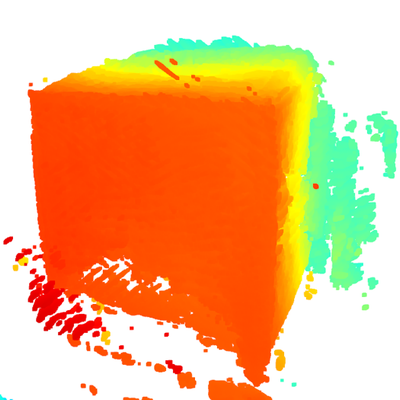}

    \vspace{2pt}    
    \includegraphics[width=0.16\linewidth]{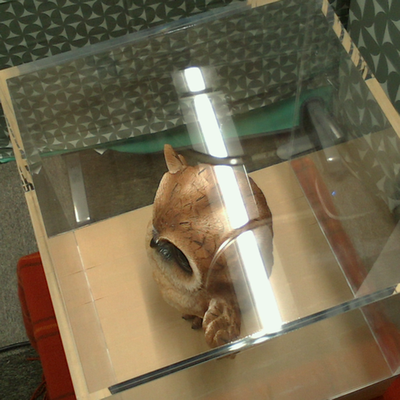}
    \includegraphics[width=0.16\linewidth]{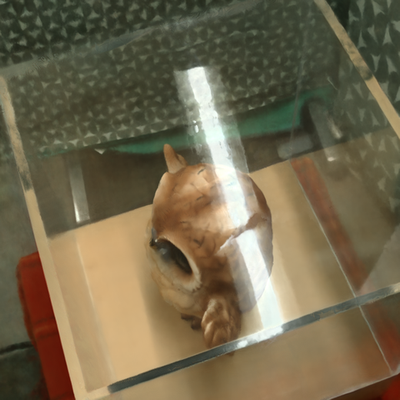}
    \includegraphics[width=0.16\linewidth]{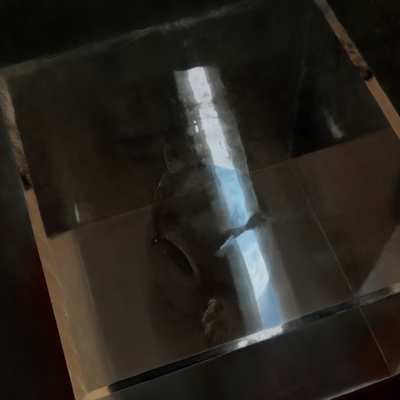}
    \includegraphics[width=0.16\linewidth]{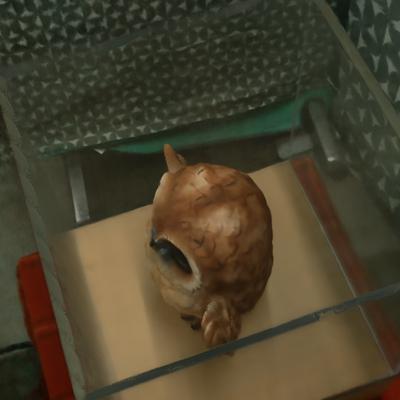}
    \includegraphics[width=0.16\linewidth]{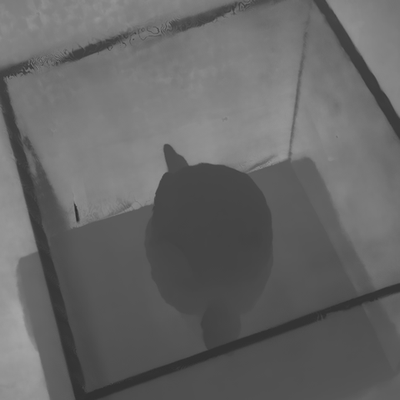}
    \includegraphics[width=0.16\linewidth]{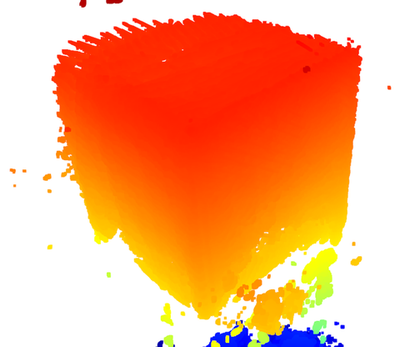}
    
    \vspace{2pt}    
    \begin{subfigure}{0.16\textwidth}
    \captionsetup{justification=centering}
    \caption{Ground truth\newline }
    \end{subfigure}
    \begin{subfigure}{0.16\textwidth}
    \captionsetup{justification=centering}
    \caption{Rendered image\newline}
    \end{subfigure}
    \begin{subfigure}{0.16\textwidth}
    \captionsetup{justification=centering}
    \caption{View-dependent component}
    \end{subfigure}
    \begin{subfigure}{0.16\textwidth}
    \captionsetup{justification=centering}
    \caption{View-independent component}
    \end{subfigure}
    \begin{subfigure}{0.16\textwidth}
    \captionsetup{justification=centering}
    \caption{View-independent depth}
    \end{subfigure}
    \begin{subfigure}{0.16\textwidth}
    \captionsetup{justification=centering}
    \caption{Refraction points (glass surface)}
    \end{subfigure}

    \caption{Result images of our proposal method modeling test datasets. The top row shows results from the simulation dataset and the bottom two rows show results from the real-world dataset.(a) Ground-truth image of scene viewed from a test pose. (b) The image rendered by the proposed method. (c) The view-dependent component and (d) view-independent component modeled by the proposed method. (e) Depth map image of the view-independent component. (f) Points where refraction greater than a threshold occurred as a point cloud. }
    \label{fig:proposal_method_result}
    
\end{figure*}

\subsection{Experimental dataset}
\subsubsection{Simulation dataset} In this study, we generated a simulation dataset using Blender~\cite{blender}. 
The experimental dataset includes several scenes containing a glass showcase with the size of $50~\text{cm} \times 50~\text{cm} \times 50~\text{cm}$ surrounded by walls with textures.
Here, the thickness of the glass is $1~\text{cm}$, and the refractive index is 1.45. 
Inside this showcase is an object in \{Lego, House, Color Ball, Flower\}. 
We also generated a dataset of Lego object placed in art gallery, which is surrounded by wall with paintings.
We placed a ceiling light with area in the scene and utilized Blender's BSDF model to render the scene with physical simulations of both the light and the glass. 
One set of a scene comprised 200 training images, and both the test and validation sets contained 25 images each with their respective intrinsic and extrinsic camera parameters. 

\subsubsection{Real world dataset} 
We developed an image-capturing system to take images using a camera, Realsense L515, and a robotic arm, UR-10e, as shown in Fig.~\ref{fig:scanning}. 
Note that though L515 is an RGB-D camera, we did not use the depth information for 3D reconstruction. 
Visual camera tracking methods struggle to function correctly when glass occupies a large portion of each image, presenting challenges in accurately estimating camera poses. 
Therefore, we used the robot arm to take the images from known camera poses\cite{wang2019autonomous, aanaes2016large}. 
We utilized 170 images, taken as uniformly as possible, covering a quarter sphere primarily near the arm for the training.
The case size is $30~\text{cm} \times 30~\text{cm} \times 30~\text{cm}$, and the thickness of the glass is $8~\text{mm}$. 

\subsection{Implementation and training}
We implemented the proposed method based on NeRF-Pytorch~\cite{lin2020nerfpytorch}. 
The training process sampled the rays corresponding to 1,024 pixels from a randomly selected training image in each iteration. 
In addition, each ray initially sampled 128 points at uniform intervals.
In the coarse-to-fine strategy of NeRF, an additional 64 points are sampled in a hierarchical manner for segments with higher densities inferred from the coarse model. 
The proposed method samples an additional 32 points using the glass density and 32 points using the view-independent density.
This results in a total of 192 sampling points being input into the fine model. 
We set the feature vector's dimension $\theta$ to 64. 
We performed a total of 200,000 training iterations on an Nvidia RTX 4080 graphics processing unit, which took approximately 10 hours for a single scene.
\begin{table*}[]
\centering
\caption{Results of each Dataset and methods (simulation dataset)}
\label{tab:comparison}
\resizebox{\textwidth}{!}
{
\tabcolsep = 1.6pt
\begin{tabular}{@{}l|ccc|ccc|ccc|ccc|ccc@{}}
\multicolumn{1}{c}{} & \multicolumn{3}{c}{Lego} &  \multicolumn{3}{c}{House}&  \multicolumn{3}{c}{Color Ball} &  \multicolumn{3}{c}{Flower}&  \multicolumn{3}{c}{Lego (Gallery)} \\ 
Method & PSNR $\uparrow$  & SSIM  $\uparrow$ & LPIPS $\downarrow$ &  PSNR $\uparrow$  & SSIM  $\uparrow$ & LPIPS $\downarrow$ &  PSNR  $\uparrow$ & SSIM $\uparrow$  & LPIPS $\downarrow$ &  PSNR $\uparrow$  & SSIM $\uparrow$  & LPIPS   $\downarrow$  &  PSNR $\uparrow$  & SSIM $\uparrow$  & LPIPS   $\downarrow$ \\ \midrule
NeRF~\cite{nerf}            & 29.0111     & 0.8893     &0.2984     & 30.9108     & 0.9029     &0.2951  & 29.2773     & 0.9054     &   0.2766   &  27.8217    & 0.8557     &  0.3351   &  31.7363    & 0.9184     &  0.2772  \\
Mip-NeRF~\cite{mipnerf}        & 28.9192 & 0.8923 & 0.2979 & 30.6958 & 0.9100 & 0.2838 & 29.9715 & 0.9211 & \textb{0.2346} & 27.9082 & 0.8711 & \textb{0.3055} & 32.1508 & 0.9267 & 0.2565 \\
Ref-NeRF~\cite{refnerf}        & 28.2375     &  0.8679    & 0.3357   & 29.6852     &  0.8836    & 0.3262  &  29.0698    &  0.8944    &  0.2880   &  27.3446    & 0.8396     &  0.3642  &  31.3128    & 0.9104     &  0.3011 \\
LB-NeRF~\cite{lbnerf}         & 29.8683 & \textb{0.9094} & \textb{0.2711} & 30.9116 & 0.9028 & 0.2980 & 30.0377 & 0.9151 & 0.2525 & 28.4501 &  0.8683 & 0.3198  & 32.4674 &  \textb{0.9324} & \textb{0.2564} \\ 
MS-NeRF~\cite{msnerf}         & \textb{31.0450} & 0.9060 & 0.2817 & \textb{33.3059} & \textb{0.9231} & \textb{0.2822} & \textb{31.6074} & \textb{0.9230} & 0.2594 & \textb{29.1936} & \textb{0.8740} &  0.3197 & \textb{33.0800} & 0.9274 &  0.2665 \\
Proposal Method & \textbfb{33.1834} & \textbfb{0.9357} &  \textbfb{0.2073} & \textbfb{35.3665} & \textbfb{0.9483} &  \textbfb{0.1929}    & \textbfb{33.0759}   & \textbfb{0.9453}     &  \textbfb{0.1842} & \textbfb{30.6254}   & \textbfb{0.8958}     &  \textbfb{0.2637}       & \textbfb{35.3476}   & \textbfb{0.9507}     &  \textbfb{0.1913}       
\end{tabular}
}
\end{table*}

\begin{table*}[]
\centering
\caption{Results of each Dataset and methods (real-world dataset)}
\label{tab:comparison_realdata}
\resizebox{\textwidth}{!}
{
\tabcolsep = 1.6pt
\begin{tabular}{@{}l|ccc|ccc|ccc|ccc|ccc@{}}
\multicolumn{1}{c}{} & \multicolumn{3}{c}{Owl (simple bg)} &  \multicolumn{3}{c}{Owl (textured bg)}&  \multicolumn{3}{c}{Owl (white bg)} &  \multicolumn{3}{c}{Lion (white bg)}&  \multicolumn{3}{c}{Dog (white bg)} \\ 
Method & PSNR $\uparrow$  & SSIM  $\uparrow$ & LPIPS $\downarrow$ &  PSNR $\uparrow$  & SSIM  $\uparrow$ & LPIPS $\downarrow$ &  PSNR  $\uparrow$ & SSIM $\uparrow$  & LPIPS $\downarrow$ &  PSNR $\uparrow$  & SSIM $\uparrow$  & LPIPS   $\downarrow$  &  PSNR $\uparrow$  & SSIM $\uparrow$  & LPIPS   $\downarrow$ \\ \midrule
LB-NeRF~\cite{lbnerf} & 26.1937 & 0.8388 & 0.4037 & 25.2232 & 0.7725 & 0.4484 & 26.1833 &  0.8649  & 0.4370   &  26.3987 & 0.8427 & 0.4562 & 26.5699  & 0.8577  & 0.4543  \\ 
MS-NeRF~\cite{msnerf} & \textbfb{27.5344} & \textbfb{0.8599} & 
\textbfb{0.3496} & \textbfb{26.8352} & \textbfb{0.8006} & \textb{0.4048} & \textbfb{28.2428} & \textbfb{0.8748} & \textbfb{0.3984} & \textb{28.5226} & \textb{0.8722} & 
\textb{0.4177} & \textbfb{27.7415} & \textbfb{0.8691} & \textbfb{0.4191} \\
Proposal Method & \textb{26.7532} & \textb{0.8541} &  \textb{0.3823} & \textb{26.6856} & \textb{0.8001} & \textbfb{0.3877} & \textb{27.4028} & \textb{0.8721} & \textb{0.4041}  & \textbfb{28.6274} & \textbfb{0.8793} & \textbfb{0.4162} & \textb{27.2331} & \textb{0.8656} & \textb{0.4336} 
\end{tabular}
}
\end{table*}


%
\subsection{Evaluation}
\label{sec:evaluation}
Figure \ref{fig:proposal_method_result} shows examples of the images obtained by the proposed method using test images and corresponding camera poses. 
Here, each column (from left to right) shows the ground-truth image (i.e., the test image), the rendered image, the view-dependent component, the view-independent component, the corresponding depth image, and the point-cloud of refraction points. 

The view-dependent images with specular reflection components demonstrate that the reflections were extracted correctly by the proposed method. 
In addition, the view-dependent image also shows that the reflection component was removed effectively. 
The depth image was estimated correctly, which indicates that the refraction was estimated well, and the reflective component, which is problematic for shape estimation, was removed effectively.

The proposed method estimates the points where refraction occurs explicitly; thus, we can evaluate the accuracy of the estimated refractive surface by using the simulation dataset to determine how close these points are to the original glass surface. 
We determined that the estimated glass surface point where the offset $\Delta x \times w$ is greater than a threshold value (0.01 cm in this experiment) when rendering test images. 
The estimated glass surface is shown in Fig.~\ref{fig:proposal_method_result} (f). 
The average errors of the estimated glass surfaces for Lego, House, Color Ball, Flower, Lego (Gallery) were $0.4657~\text{cm}$, $0.4728~\text{cm}$, $0.3553~\text{cm}$, $0.2693~\text{cm}$, $0.9662~\text{cm}$, respectively. 
Although there are some outliers, it can be seen that reasonably good results were obtained by the proposed method.


%
%
%
%
%

\subsection{Comparative evaluation}
In this evaluation, the proposed method was compared with several NeRF-based methods. 
Here, we applied the original NeRF~\cite{nerf}, Mip-NeRF~\cite{mipnerf}, Ref-NeRF~\cite{refnerf}, LB-NeRF~\cite{lbnerf}, and MS-NeRF~\cite{msnerf} methods to the 
simulation dataset constructed in this study. 
Note that no open source code is available for LB-NeRF; thus, we implemented its structure by adding a 3D offset estimated from a concatenation of the 3D point's positions and view direction using a fully-connected neural network with seven layers containing 256 nodes. 
In this evaluation, evaluation metrics commonly used for image comparison were used to assess the compared methods, i.e., the peak signal-to-noise Ratio (PSNR), the structural similarity index measure (SSIM)~\cite{wang2004image}, and the learned perceptual image patch similarity (LPIPS)~\cite{zhang2018unreasonable}. 

Table \ref{tab:comparison} shows the results on the simulation dataset. 
As can be seen, the proposed method outperformed all compared methods on each dataset. 
Among the compared method, the MS-NeRF method obtained comparatively superior results. 
Note that the MS-NeRF method does not clearly determine the decomposed spaces for nonmirror surfaces; thus, it enhances the image synthesis precision by separating the intense reflective components associated with glass.

\color{black}
Table~\ref{tab:comparison_realdata} shows the results on the real-world dataset. 
Based on the results of the simulation data, Table~\ref{tab:comparison_realdata} contains only comparisons with LB-NeRF and MS-NeRF. 
In the context of novel view synthesis, the proposed method and MS-NeRF had similar scores. 
Due to the difference in the base method, the performance is slightly different depending on the background, illumination, and target object.
However, as mentioned above, the proposed method is superior to other methods because it can separate the reflection components and estimate the refraction position and magnitude. 
\color{black}


\section{Conclusion}

In this paper, we have proposed a method that utilizes implicit neural representations to model scenes with objects enclosed in a glass case.
The proposed method distinguishes between the refraction and reflection effects by learning them with view-independent and view-dependent components, and by determining the refraction points indicative of the glass surfaces simultaneously.
The proposed method was evaluated experimentally, and the experimental results demonstrate that the proposed method is proficient in terms of separating the components that vary with viewpoint from those that do not. 
On the other hand, as explained in Sec. \ref{sec:evaluation}, the separation of view-dependent and view-independent components may not be perfect. 
Improving the accuracy of component separation is one of the future works.

\bibliographystyle{IEEEtran}
\bibliography{IEEEfull,myref-s} 

\end{document}